\documentclass[letterpaper, 10 pt, conference]{ieeeconf}  
\usepackage{algorithmic}
\usepackage{epsfig}
\usepackage{graphicx}
\usepackage{amsmath}
\usepackage{amssymb}
\usepackage{booktabs}
\usepackage{multirow}
\usepackage{multicol}
\usepackage[ruled,vlined]{algorithm2e}
\usepackage{xcolor}
\usepackage{pifont}
\usepackage{csquotes}

\IEEEoverridecommandlockouts                              

\usepackage{booktabs}                                   
\usepackage{multirow}                                   
\usepackage{makecell}                                   
\usepackage{tablefootnote}                              
\usepackage[symbol]{footmisc}                           
\usepackage{amsmath,amssymb}                            
\usepackage{xcolor}                                     
\let\labelindent\relax              
\usepackage{enumitem}                                   
\usepackage{subcaption}                                 
\usepackage{stfloats}                                   
\usepackage[misc]{ifsym}     
\usepackage{hyperref}   

\newcommand{\eat}[1]{}                                  

\title{\LARGE \bf
Language-Conditioned Robotic Manipulation \\
with Fast and Slow Thinking
}

\author{Minjie Zhu$^{1,*}$, Yichen Zhu$^{2,*}$, Jinming Li$^{3}$, Junjie Wen$^{1}$, Zhiyuan Xu$^{2}$, Zhengping Che$^{2}$, \\
Chaomin Shen$^{1}$, Yaxin Peng$^{3,\dagger}$, Dong Liu$^{2}$, Feifei Feng$^{2}$, and Jian Tang$^{2,\dagger}$
\thanks{$^{1}$School of Computer Science, East China Normal University, China
        {\tt\small \{51255901028, 51255901019\}@stu.ecnu.edu.cn, cmshen@cs.ecnu.edu.cn}}
\thanks{$^{2}$Midea Group, China
        {\tt\small \{zhuyc25, xuzy70, chezp, liudong13, feifei.feng, tangjian22\}@midea.com}}
\thanks{$^{3}$Department of Mathematics, School of Science, Shanghai University, China
        {\tt\small \{ljm2022, yaxin.peng\}@shu.edu.cn}}
\thanks{
    $^*$Equal contributions. This work was done during Minjie Zhu, Jinming Li, and Junjie Wen's internship at Midea Group.
}
\thanks{
    $\dagger$Corresponding authors: Yaxin Peng and Jian Tang.
}
}

\begin{document}

\maketitle
\thispagestyle{empty}
\pagestyle{empty}

\begin{abstract}
The language-conditioned robotic manipulation aims to transfer natural language instructions into executable actions, from simple \enquote{pick-and-place} to tasks requiring intent recognition and visual reasoning. Inspired by the dual-process theory in cognitive science—which suggests two parallel systems of fast and slow thinking in human decision-making—we introduce \textit{Robotics with Fast and Slow Thinking (RFST)}, a framework that mimics human cognitive architecture to classify tasks and makes decisions on two systems based on instruction types. Our RFST consists of two key components: 1) an instruction discriminator to determine which system should be activated based on the current user's instruction, and 2) a slow-thinking system that is comprised of a fine-tuned vision-language model aligned with the policy networks, which allow the robot to recognize user's intention or perform reasoning tasks. To assess our methodology, we built a dataset featuring real-world trajectories, capturing actions ranging from spontaneous impulses to tasks requiring deliberate contemplation. Our results, both in simulation and real-world scenarios, confirm that our approach adeptly manages intricate tasks that demand intent recognition and reasoning. The project is available at \href{https://jlm-z.github.io/RSFT/}{https://jlm-z.github.io/RSFT/}.
\end{abstract}

\section{Introduction}
Originally designed to generate robot actions based on language instructions and observations, robotic controls have demonstrated an expanding capability to handle a broader array of manipulation tasks beyond simple pick-and-place operations. Interestingly, at the heart of this manipulation model lies an auto-regressive mechanism for trajectory generation, which offers a direct mapping from an \enquote{instruction-observation} pair to an action space~\cite{brohan2023rt,jang2022bc}. Can such a straightforward mechanism truly be the foundation for a robot aiming to become a general agent, assisting humans in real-world scenarios? If it falls short, what issues might challenge this current approach, and what alternative mechanisms should be considered?

The literature concerning human cognition offers insights into these questions. Dual-process model research indicates that individuals engage with decisions in two primary ways: a rapid, instinctive, subconscious manner (referred to as \enquote{System 1 or Fast-thinking}) and a measured, deliberate, conscious manner (\enquote{System 2 or Slow-thinking})~\cite{causalrep, stanovich1999rational, kahneman2002representativeness, sloman1996empirical, kahneman2011thinking}. Notably, these two systems have been associated with various mathematical models in machine learning. For instance, studies on reinforcement learning in both humans and animals have delved into conditions that prompt either associative \enquote{model-free} learning or the more contemplative \enquote{model-based} planning~\cite{scholkopf2021toward, bubeck2023sparks, moerland2023model}. The straightforward associative command-action of the policy network bears similarities to \enquote{System 1}. Therefore, it could be enhanced with a more intentional \enquote{System 2} planning approach. This would involve reasoning that (1) preserves and examines a range of options for present choices beyond the straightforward command like \enquote{pick-and-place objects} and (2) assesses its existing state, proactively forecasting or revisiting decisions for a more comprehensive perspective.

\begin{figure}[t]
    \centering
    \includegraphics[width=0.47\textwidth]{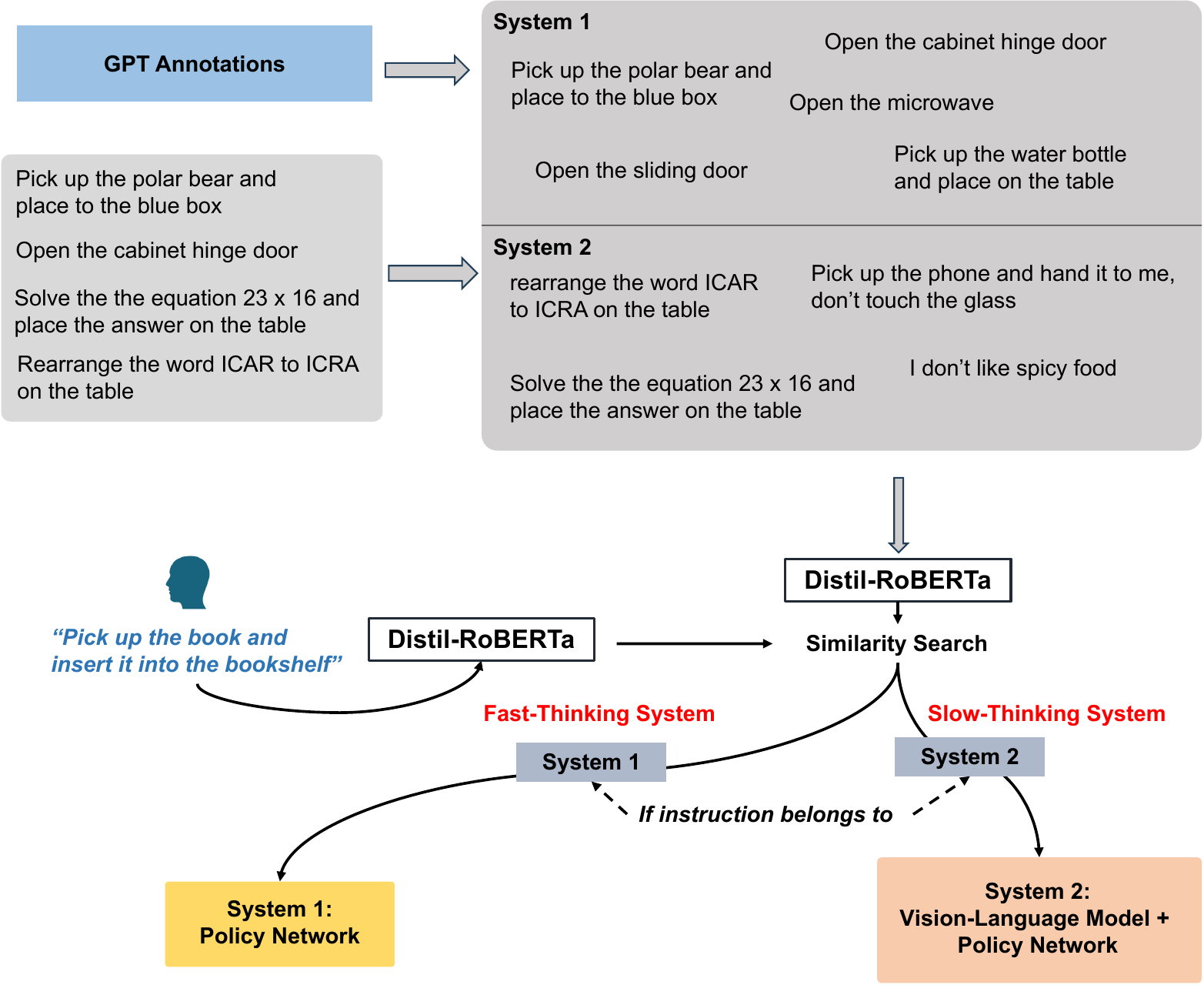}
    \caption{The overview of RFST. We collected a number of instructions and employed GPT4~\cite{gpt4} for annotation. Upon receiving an instruction, the robot processes it through Distil-RoBERTa to obtain an embedding. Leveraging embedding similarity search, we classified the instruction into either a fast-thinking system or a slow-thinking system.}
    \label{fig:fst_summary}
\end{figure}

To design such a planning process, we draw inspiration from the human cognitive system, originated from Kah~\cite{kahneman2011thinking}. We propose a novel language-conditioned Robotic manipulation framework with Fast and Slow Thinking (RFST, in short), depending on the complexity of the user's language instruction. As Figure~\ref{fig:fst_summary} illustrates, while existing methods simply output the robot's action via a policy network, we actively maintain a Think Bank, where each thought is divided into either a fast-thinking system or a slow-thinking system that serves as an intermediate step toward problem-solving. Such a high-level semantic unit allows the robot to self-evaluate the progress different thoughts make toward solving the problem through a deliberate reasoning process or an intuitive action. 
Finally, we combine this language-conditioned capability to perform manipulation tasks. We leverage different models for two types of systems. As System 1 only involves fast and straightforward thinking, we allow a simple, shallow policy network to do the jobs. For difficult tasks that need reasoning or planning, we opt for a Vision-Language Model (VLM). This model is designed to either break down the tasks into manageable sub-tasks or clarify the user's intent. Subsequently, a policy network outputs action based on these augmented instructions.

Empirically, we validate the efficacy of RFST across a spectrum of tasks, ranging from basic pick-and-place and rotation to more intricate tasks such as mathematical and visual reasoning. While traditional robotic manipulation methods can address the former tasks, the latter requires systematic planning or an in-depth search for the true user intention, challenges that direct policy networks often struggle with. Our results indicate that RFST delivers superior performance on complex tasks in simulated benchmarks. Moreover, we have curated a dataset featuring real-world trajectories spanning nine tasks: three for fast-thinking systems and six for slow-thinking systems. Our data reveals that RFST can adeptly tackle tasks from both categories. 

\noindent
\textbf{To sum up, our contributions are threefold:}
\begin{itemize}
    \item We present a fast and slow thinking framework for robotics manipulation that categorizes incoming instruction into two systems and performs control correspondingly.
    \item We design a framework for slow thinking, which leverages the fine-tuned VLM to perform instruction-observation conditioned reasoning and re-write the instruction for robotics affordance. 
    \item We collect a set of real-world datasets, including tasks like math reasoning and intent recognition, and examine the effectiveness of our approach on both simulation and real-world scenarios. 
\end{itemize}

\section{Related Work}
\noindent
\textbf{Reasoning in Language and Vision.} Chain-of-thought~\cite{cot} use was a coherent language sequence that served as a meaningful intermediate step toward problem-solving of mapping input questions with output language. Self-consistency with CoT~\cite{cotsc} ensembles CoT and prioritizes the most frequent output. Tree-of-Thought~\cite{tot} uses a tree structure to classify input questions into different sub-trees for the final answer. Least-to-most prompting~\cite{zhou2022least} breaks down a complex problem into a series of simpler subproblems and then solves them in sequence. Program-of-Thought~\cite{pot} translates natural language into program format, assisting LLM in mathematical reasoning. Collectively, these \enquote{X-of-Thought} methodologies empower LLM to engage in chats that demand reasoning.

A recent trend in vision-language models~\cite{li2023blip-2, liu2023llava, clip, shridhar2022cliport, LLaVA-phi} allows for the comprehension of images and the provision of answers in natural language, albeit with constrained reasoning capabilities. These reasoning skills are realized by integrating large language models~\cite{llama} with a vision backbone. Instead, we establish a framework that discerns the \enquote{amount of minds} needed to process the instruction.


\noindent
\textbf{Large Language Model for Robotics.} Large language models possess the power of reasoning. With the advancement of LLM in the past year, a rising number of projects have been proposed to use LLM as a high-level model for task planing~\cite{fan2022minedojo}, code generation~\cite{cop}, navigation~\cite{mandi2023roco,visualmapnav}, and action correction~\cite{cui2023no, oci}. RT-2~\cite{brohan2023rt} introduced an end-to-end model capable of processing tasks with text and one or more images, producing a sequence of tokens to control a robot. When integrated with expansive vision-language models like PaLI-X~\cite{chen2023pali} and PaLM-E~\cite{driess2023palm}, RT-2 demonstrates reasoning within this framework. While the end-to-end method is indeed appealing, nevertheless, these networks typically demand vast amounts of training data and come with considerable computational costs. In contrast, our proposed RFST allows reasoning, symbolic understanding, and intent recognition. It maintains the delicate balance of high-level reasoning with low-level control models. Furthermore, RFST requires significantly less training data, making it advantageous when paired with extensive datasets.

\section{Methodology}
In this section, we provide a detailed description of RFST. We first give an informal definition of fast and slow thinking. Then, we introduce the overall framework of RFST and present our slow-thinking system.

\subsection{Formal Definition of Fast Thinking and Slow Thinking}
Given a language instruction $x$, the policy network is a mapping function to get an output $y$. 
The complexity of the mapping function $p_{\theta}$ is determined by the $x$. For a simple instruction, e.g., pick up an apple, the mapping function could be simple $y \sim p_{\theta}(x)$. We consider these tasks as fast-thinking tasks. When the mapping of input $x$ to output $y$ is non-trivial (i.e., when x is a math question and y is the numerical answer), we need to introduce an intermediate step to $z$ to bridge $x$ and $y$. Then, the mapping function is $y \sim p_{\theta}(x|z)$. Our task is to classify the given instruction as either a fast-thinking system or a slow-thinking system. Notice that the fast-thinking system can be arbitrary language-conditioned robot manipulation algorithm that has been well developed over the past year, i.e., GATO~\cite{gato}, VIMA~\cite{jiang2022vima}, RT-1~\cite{brohan2023rt}. Therefore, in the second part, we focus on introducing our slow-thinking system, which we need to design the $z$ for correct manipulation meticulously.

\subsection{Overall Framework of RFST}
To determine whether an incoming language instruction corresponds to System 1 or System 2, we have established an instruction bank comprising many language instructions. We employ GPT-4~\cite{gpt4} to simulate a robot. Given specific scenarios, we prompt GPT-4 to produce a list of language instructions and specify their association with either System 1 or System 2. Using this curated set of categorized instructions as our foundational seed, we enable GPT-4 to augment these instructions, reshaping them into diverse formats while maintaining consistent meaning. After ten iterations, this process yields thousands of pre-classified language instructions. Additionally, we go through a manual review of the generated instructions for accuracy. We give an overview in Figure~\ref{fig:fst_summary}.

Because the instruction is entirely generated by the Large Language Model (LLM), it is obvious that LLM can be used to decide which category the user utterances. However, due to the significant computational demands of LLM, we have been motivated to seek a more lightweight approach. Toward this goal, we encode the language and undertake instruction retrieval. The utterance is embedded using a frozen version of the Distil-RoBERTa~\cite{sanh2019distilbert, liu2019roberta} language model, as provided by the Sentence-BERT~\cite{reimers2019sentence} project. Supported by an \enquote{unnatural language processing} nearest neighbors index, inference-time utterances are matched with the closest training exemplars. These exemplars are then retrieved and processed by the model. The classification of a given instruction is determined based on the category of the retrieved instruction. Empirically, we used 500 instructions from GPT-4 to form our \enquote{Think Bank} and do instruction classification at test time. In our experiments, this approach can perfectly classify language instruction.

\begin{figure}
\centering
\includegraphics[width=0.9\linewidth]{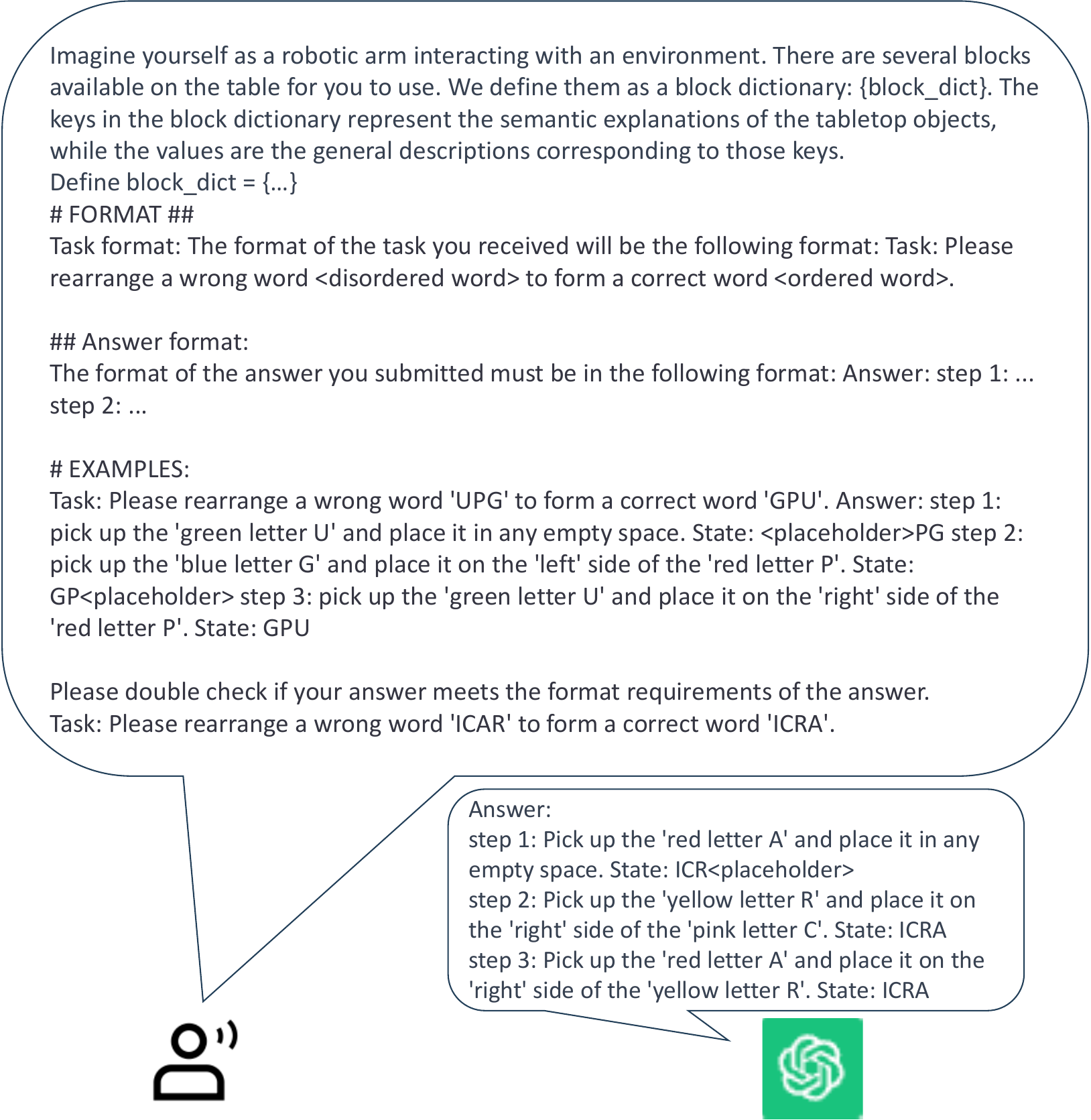}
\caption{An illustrative example of step-by-step task planning originates from GPT-3.5-turbo. The planning produced by the LLM serves as the foundation for formulating our text-image pairs used for VLM training.}
\label{fig:gpt}
\end{figure}

\subsection{Details of Slow Thinking}
We illustrate the details of our slow-thinking mode. There are two key factors in System 2: 1) a vision-language model that could perform reasoning and intent recognition, given the language instruction and current observation, and 2) a policy network that can understand the planning from the vision-language model to act precisely.

\noindent
\textbf{Empower Vision-Language Model with Reasoning and Intent Recognition.} 
The vision-language models we utilize in this study accept a text-image pair as input and yield a sequence of tokens, typically representing natural language text. These models are versatile, capable of a broad spectrum of visual interpretation tasks—from deciphering an image's composition to responding to queries about individual objects and their relationships. However, standard pre-trained vision-language models lack an understanding of the physical world. Our objective is a vision-language model that not only grasps the relationship between observed scenes and natural language, but can also recognize the user's intention and provide logical, step-by-step instructions to guide a robot's actions. To realize this, we require a dataset featuring instruction-observation pairs and the finetune a VLM.

\begin{figure}
\centering
\includegraphics[width=\linewidth]{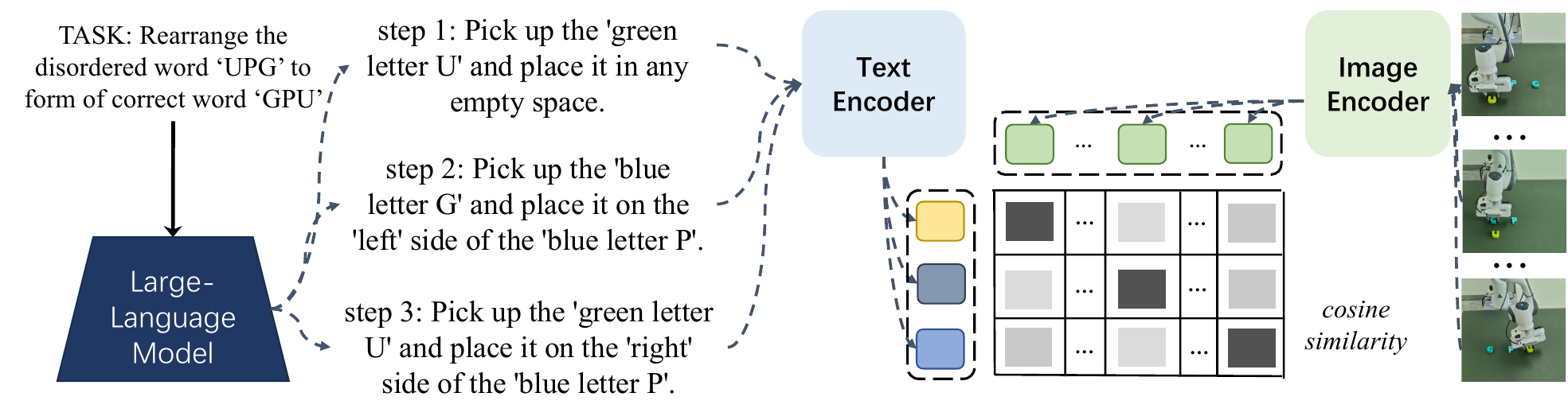}
\caption{An illustration of CLIP computing the similarity between step-wise text description and observations.}
\label{fig:clip}
\end{figure}

\noindent
\textbf{Multi-modal Planning Data Collection.} We demonstrate how to build up our multi-modal instruction data. First of all, we need to delineate tasks step-by-step, aligned with the user's intention. To achieve this, we seek the help of a Large Language Model (LLM). We first convert the scene into natural language to ensure the LLM comprehends it effectively. We use the pre-trained vision-language model, i.e., BLIP-2~\cite{li2023blip-2}, to do the image caption. Then, for each set of tasks, such as math reasoning, grammar check, and user's intention understanding, we draft a prompt script. This script incorporates in-context learning and a chain-of-thought approach, enabling the LLM to yield our anticipated planning or clarify user intentions. We give an example of the prompt for instruction generation of word reordering is present in Figure~\ref{fig:gpt}. After gathering all the data, manual verification was conducted. We also include a number of texts that recognize the user's intention and transfer them into actionable instructions for the robot. All these text scripts are used for training only, and they are generated by GPT-4. Empirically, we found the majority of responses from GPT-4 were accurate.

\noindent
\textbf{Mapping Sub-goal with Observations.} For tasks requiring step-by-step planning, a conventional approach involves using these steps as instructions and subsequently grounding them into robotic actions. Yet, to ensure the robotic agent thoroughly understands the instructions, it's crucial to synchronize the instruction with the observation for that specific step. We advocate the use of CLIP~\cite{clip} to bridge visual inputs with text descriptions. By computing the dot product between the text and image embedding vectors, we pair a text and image if the result surpasses a threshold value, denoted as $\alpha$. In our implementation, $\alpha$ is set to 0.75. Furthermore, we fine-tune CLIP using a limited dataset from the scene, which we've labeled manually. A brief illustration can be found in Figure~\ref{fig:clip}. To ensure accuracy, we manually inspect the data post-processing. Unlike the planning and user intention understanding derived from GPT-4 in the preceding step, manual verification is vital since CLIP's accuracy can waver if observations between two consecutive steps aren't sufficiently distinct.

\noindent
\textbf{Vision-Language Model Architecture.} We employed the pre-trained ViT-L/14 from CLIP as our visual encoder, paired with the LLaMA-2-7B as our LLM~\cite{clip, llama, llama-2}. To maintain modal alignment and facilitate a compatible input dimension for the LLM, a fully connected layer has been integrated. This layer transforms the ViT's output embedding $16 \times 16$ output embedding $V \in \mathbb R^{16 \times 16 \times 1024}$ to $V^{'} \in \mathbb R^{256 \times 4096}$. We tap into the robust vision-language capabilities inherent to the text-image alignment~\cite{zhu2023minigpt}. Moreover, we fine-tune the associated networks, holding the language and visual embeddings constant. Only the alignment layers are subject to adjustments.

\noindent
\textbf{Policy Networks with Language Instruction.} To craft an efficient multi-task robotic policy, we utilize policy networks featuring a multi-task decoder architecture. Specifically, our goal is to derive a robotic policy represented by $\pi(a_{t}|P, H)$, where $H := \{o_{1}, a_{1}, o_{2}, a_{2}, \cdots, o_{t}\}$ encapsulates the trajectory of historical interactions. The $o_{t} \in \mathbb O$ and $a_{t} \in \mathbb A$ denote the observations and actions at each interaction step, respectively. These policy networks are designed to handle multi-modal tokens, and for their encoding, we incorporate multi-modal prompts. The images are processed via a vision backbone (ResNet-50~\cite{nair2022r3m, he2016deep}) while the text is tokenized. The image embedding and text embedding are connected with FiLM~\cite{perez2018film}. The policy network is followed up, and it consists of three MLP layers with ReLU activation.

\begin{figure*}
\centering
\includegraphics[width=0.75\linewidth]{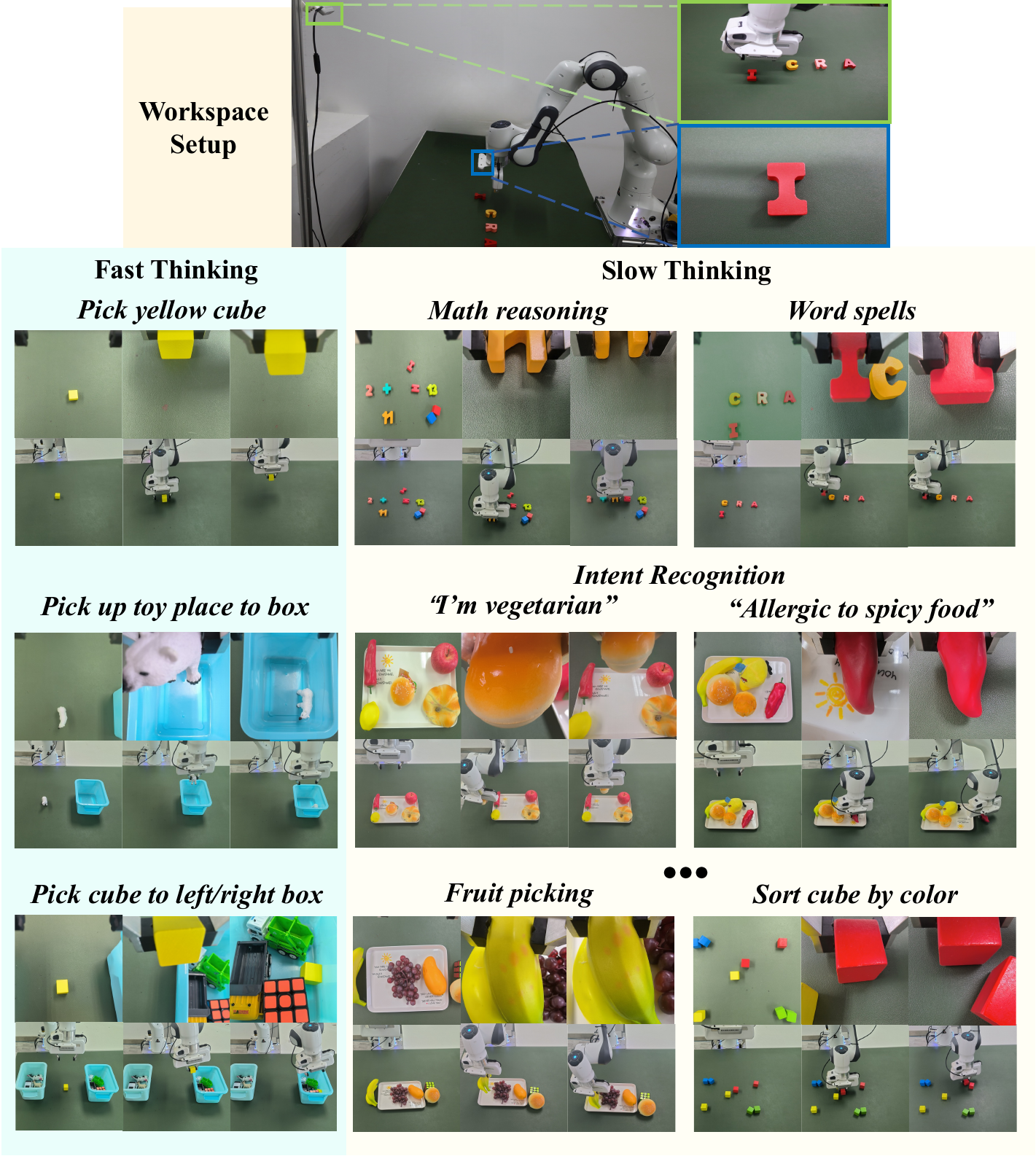}
\caption{We collect a dataset with real-world trajectories using a Franka robotic arm. Each trajectory is a sequence of images from two cameras. We consider multiple tasks that belong to either the fast-thinking system or the slow-thinking system.}
\label{fig:dataset}
\end{figure*}
\section{A Dataset of Two System}
To study our pre-training approach, we collect a large dataset of real-world robot trajectories. We collect different tasks that belong to different systems. In this section, we describe our data collection process and show qualitative examples from our dataset. 

\noindent
\textbf{Hardware.} We use the Franka robot with a 7-degree-of-freedom arm, which is equipped with a parallel jaw gripper (see Figure~\ref{fig:dataset}, top). Proprioceptive data, including joint positions and velocities, are recorded throughout our experiments. Actions in the joint space are determined by the differences between successive states. Our workspace boasts two high-quality D435i RealSense RGBD cameras. We only use the RGB information in our experiments. One egocentric camera is attached to the robot's hand, and one exocentric camera is positioned at the robot's front.

\noindent
\textbf{Math Reasoning [Slow-Thinking].} Our objective is to engage the robot in mathematical reasoning tasks, including equation solving. We present two sets of tasks. The first requires the robot to directly compute the mathematical equation presented on a table. The second involves solving for an unknown variable $x$. For example, when presented with an image displaying $11 \times 13 = $, or $1 + x = 6$, the robot is tasked with completing the equation or substituting $x$ with the correct number. These tasks are generally single-step challenges. Their success hinges on the vision-language model's capability to comprehend the mathematical reasoning within the scene.

\noindent
\textbf{Word Correction [Slow-Thinking].} The robot is responsible for correcting word spellings, be it due to incorrect sequences or specific word designations. These tasks can range from simple single-step actions to more intricate multi-step processes. Take, for instance, the task of rearranging the word \enquote{ICAR} to form \enquote{ICRA}. This task demands three distinct steps: 1) pick up the word \enquote{A} and place it in the empty space, 2) pick up the word \enquote{R} and place it next to the word \enquote{C}, 3) pick up the word \enquote{A} and place next to the word \enquote{R}. This kind of task not only tests the robot's linguistic aptitude but also its dexterity and ability to perform sequential operations accurately. The combination of language and motor skills is paramount to execute such tasks efficiently. 


\noindent
\textbf{Sort Cube by Color [Slow-Thinking].} We've arranged several cubes on the table, each coming in one of four distinct colors. The robot's task is to identify individual cubes, grasp them, and then group them with other cubes of the same color. The complexity of the task stems not just from the robot's ability to recognize colors but also from its spatial reasoning in determining where to place each cube to create organized color clusters. This challenges the robot's visual processing capabilities and its precision in handling objects.

\noindent
\textbf{Intent Recognition [Slow-Thinking].} We've designed several tasks that necessitate visual reasoning. Consider a scenario where an image depicts various foods on a table. If a user provides the instruction, "I'm allergic to spicy food," the robot would identify spicy items, such as chili or other spicy ingredients, and relocate them to the opposite side of the table. This exemplifies a typical situation in which robots must discern the user's intent based on verbal directives.

\noindent
\textbf{Pick Cube based on Color [Fast-Thinking].} The robot's assignment is to grasp a cube according to the color information from language instruction. 

\noindent
\textbf{Pick Cube and Place to left/right box[Fast-Thinking].} The robot is asked to select a cube by color and put it into a box, either on the left side or right side, based on the instructions.

\noindent
\textbf{Pick Toy and Place to box [Fast-Thinking].} The robot is asked to pick up a toy put into a box. 

\noindent
\textbf{Statistics for Data Collection.} We collect approximately 2,000 real-world trajectories, where the average length of the trajectories is around 100. The dataset contains variations in object poses, shape, and appearance. Objects are randomly placed on the table. We give multiple examples for our aforementioned tasks in Figure~\ref{fig:dataset}.

\section{Experiments}
In this section, we empirically assess the broad applicability of RFST across diverse tasks in both simulated and real-world settings.

\subsection{Simulation Experiments}
\noindent
\textbf{Experiments Setup.} We conduct our simulation experiments on VIMA-Bench~\cite{jiang2022vima}, built on the Ravens simulator~\cite{zeng2021transporter}. We chose VIMA-Bench because this benchmark consists of tasks that require reasoning and multi-step manipulation. We train the policy network on all six tasks. We select two tasks as fast-thinking tasks: rotation and simple object manipulation. In the \enquote{Rotation} task, the robot is instructed to rotate objects clockwise by specific degrees along the z-axis. The \enquote{Simple Object Manipulation} task involves placing one object inside another. Both these tasks are executed in a single step.

\begin{figure*}
\centering
\includegraphics[width=0.9\linewidth]{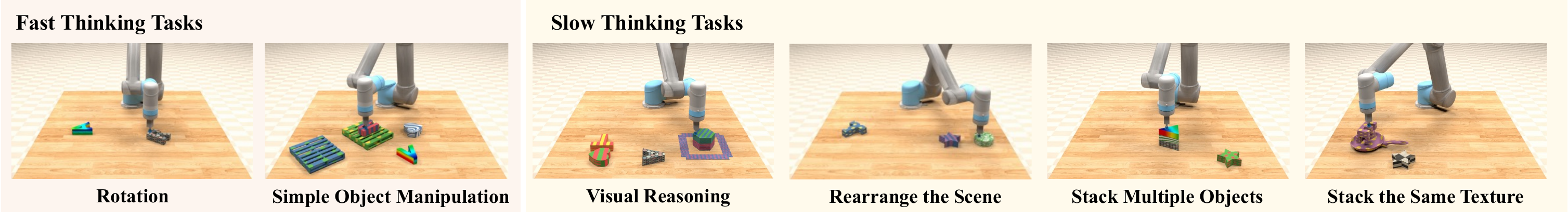}
\caption{Example of tasks in simulation. We select six tasks in VIMA-Bench~\cite{jiang2022vima} and categorize them into fast-thinking and slow-thinking tasks accordingly.}
\label{fig:vima_data_example}
\end{figure*}

For more deliberate (slow-thinking) tasks, we selected four distinct assignments. The \enquote{Rearrange the Scene} task provides a description of the desired scene, instructing the robot to rearrange objects accordingly. \enquote{Visual Reasoning} requires stacking multiple objects in a specified square sequence. The \enquote{Stacking Multiple Objects} task delineates the stacking order, allowing the model to determine the sequence of object placement. Finally, the \enquote{Stack the Same Texture} task demands the model to identify and stack objects that share identical textures. A visual representation of these six tasks can be found in Figure~\ref{fig:vima_data_example}. In our experiments, we use Task 1 and Task 2 to represent the first two fast-thinking tasks and Task 3-6 to denote the latter four slow-thinking tasks. For those slow-thinking tasks, we follow our proposed pipeline to use a trained vision-language model to give a step-by-step plan. and then feed it into our policy networks. Please note that in the standard VIMA-Bench, the instruction comprises both image and text components. For images that depict the entire scene, we employ LLaVA-13B~\cite{liu2023llava} to describe the scene and manually correct any errors. For images representing specific objects, we utilize the look-up table from VIMA-Bench to convert them into text. We train our model with 2M parameters. The number of training trajectories for each task is 1,000. We mixed rotation and pick-and-place task data to train a fast-thinking policy network. Each task is evaluated with 20 trials, where objects are randomly placed. 

\noindent
\textbf{Baselines}. We compare our model with GATO~\cite{gato}, Flamingo~\cite{alayrac2022flamingo}, and VIMA~\cite{jiang2022vima}. Gato is a decoder-only generalist agent. Flamingo is a state-of-the-art vision-language model. VIMA is a multi-task robotic manipulation model that receives multi-modal prompts. We follow the publicly released code in VIMA to implement their methods. For fair comparisons, we use text-only prompts in all experiments. All methods are trained on the same amount of data. 
\begin{table}[t]
\caption{Success rates on VIMA-Bench over six tasks. The Tasks 1 and 2 belong to fast-thinking system, and Task 3-6 belong to slow-thinking system. Our proposed RFST significantly outperforms other methods in accomplishing slow-thinking tasks, achieving notably higher success rates.}
\centering
\resizebox{0.98\columnwidth}{!}{\begin{tabular}{lcccccc}
\toprule
Method & Task 1 & Task 2 & Task 3 & Task 4 & Task 5 & Task 6 \\
\midrule
Gato & 37 & 50 & 11   & 18  & 22  & 16\\
Flamingo & 39 & 37 & 15  & 25  & 34 & 14 \\
VIMA & 58 & \textbf{52} & 27 & 17 & 31& 26\\
RFST (Ours) & \textbf{60} & 49 & \textbf{36} & \textbf{47} & \textbf{42} & \textbf{35}\\
\bottomrule
\end{tabular}}
\label{table:vima}
\end{table}

\begin{figure}[t]
    \centering
    \includegraphics[width=0.9\columnwidth]{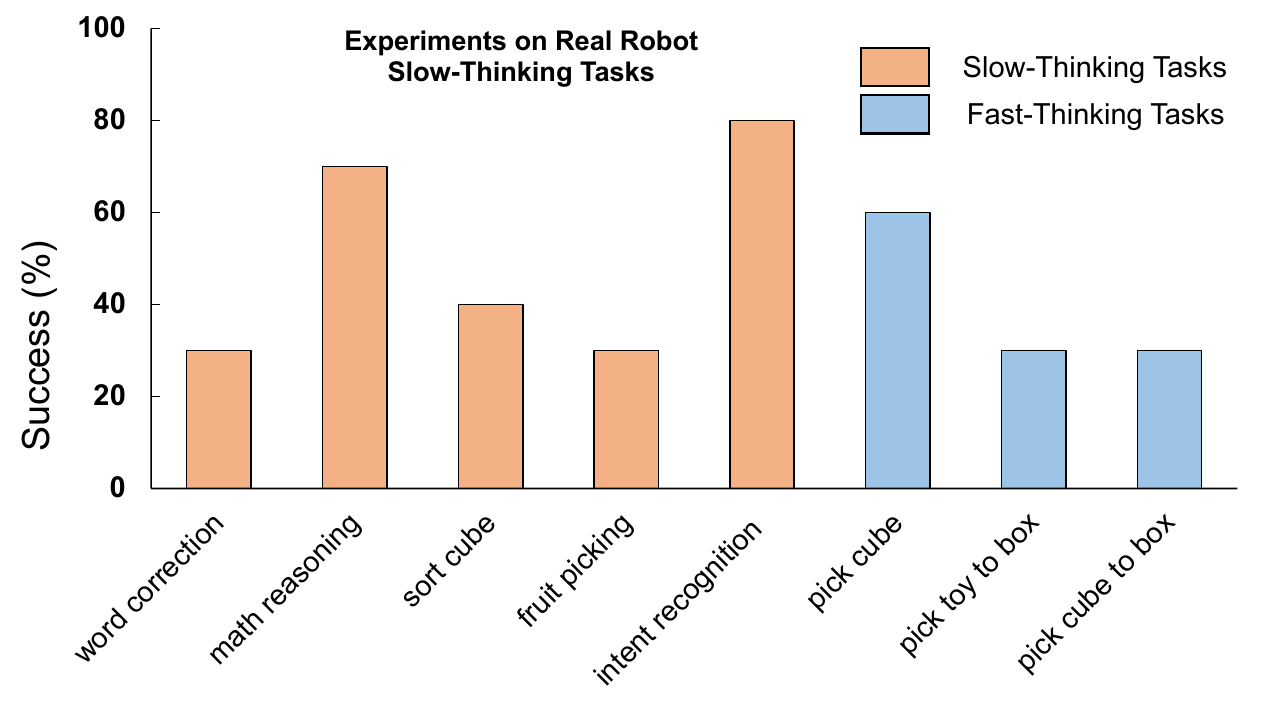}
    \caption{The experiments on the real robot. \textbf{Orange Bars}: Slow-thinking tasks. \textbf{Blue Bars}: Fast-thinking tasks. RFST empowers real robots to execute complex tasks such as mathematical reasoning and intent recognition, which were traditionally beyond the scope of conventional robotic manipulation techniques.}\label{fig:real_robot_exp}
\end{figure}

\noindent
\textbf{Main Experimental Results.} Table~\ref{table:vima} presents the experimental results. It can be observed that our model does exhibit superiority in fast-thinking tasks. For example, while we surpass VIMA by 2\% on Task 1, we lag behind by 3\% on Task 2. However, upon examining slow-thinking tasks, our approach demonstrates notably superior performance compared to the baseline. These outcomes underscore the importance of devising an appropriate intermediate step $z$ for addressing complex tasks. Such tasks may require capabilities like reasoning and symbolic understanding.

\subsection{Real-world Experiments} 
We conduct experiments on real-world robotics manipulation tasks to give a full comprehension of RFST on intent recognition, reasoning, and symbolic understanding. The real-world experiments are more challenging due to imperfect camera sensors and increased object quantity and diversity. For each task, we conduct ten trials, and the objects are randomly placed on the table. 

\noindent
\textbf{Experimental Results.} Figure~\ref{fig:real_robot_exp} demonstrates the experimental results from the real-world experiments.  On the right are the tasks that involve slow thinking. Our proposed RFST achieves good performance, especially on two reasoning tasks: math reasoning and intent recognition. The relatively low success rate of word correction is due to the size of a word being too small and non-regular, which makes it hard for the gripper to grasp it successfully. It is worth noting that our framework succeeded in intent recognition on eight out of ten trials, underscoring its exceptional capability in handling intricate tasks demanding human-like cognition.

\section{Conclusion}
We introduce an approach to robotic manipulation that seamlessly addresses both straightforward tasks, such as \enquote{pick and place}, and intricate tasks demanding visual reasoning, all within a unified framework. Our strategy draws inspiration from cognitive science, emphasizing the dual system humans employ: \enquote{System 1} for rapid, intuitive actions and \enquote{System 2} for more deliberate, contemplative thinking. To operationalize this, we have crafted tools that encompass instruction classification — determining which system an incoming instruction pertains to — and refining a vision-language model for visual reasoning. This aids policy networks in executing multi-step manipulations. Our method's efficacy is demonstrated in simulations requiring multi-step control and actual robots executing a gamut of tasks, from basic to complex. 


\bibliographystyle{IEEEtran}
\bibliography{reference}

\end{document}